\pdfoutput=1

\documentclass[11pt]{article}

\usepackage{EMNLP2022}

\usepackage{thmtools, thm-restate}
\usepackage{times}
\usepackage{latexsym}
\usepackage{bbm}
\usepackage{amsmath}
\usepackage{amssymb}
\usepackage{tabularx}
\usepackage{multirow}
\usepackage{graphicx}
\usepackage{algorithm}
\usepackage[noend]{algpseudocode}
\usepackage{linegoal}
\usepackage{multirow}
\usepackage{setspace}

\declaretheorem{proposition}

\makeatletter
\def\BState{\State\hskip-\ALG@thistlm}
\makeatother

\usepackage{arydshln}

\usepackage[T1]{fontenc}

\usepackage[utf8]{inputenc}

\usepackage{microtype}

\usepackage{inconsolata}

\title{Leveraging Open Data and Task Augmentation to Automated Behavioral Coding of Psychotherapy Conversations in Low-Resource Scenarios}

\author{Zhuohao Chen$^{1}$, Nikolaos Flemotomos$^1$\thanks{\: Work done while Nikolaos Flemotomos was at University of Southern California in 2022, and he is now affiliated to Apple Inc.}\:, Zac E. Imel$^2$, David C. Atkins$^3$, \\ \textbf{Shrikanth Narayanan$^1$}\\
  $^1$University of Southern California, Los Angeles, CA, USA \\
  $^2$University of Utah, Salt Lake City, UT, USA \\  
  $^3$University of Washington, Seattle, WA, USA \\
  {\tt sail.usc.edu  \tt zac.imel@utah.edu \tt datkins@u.washington.edu } \\}

\begin{document}
\maketitle
\begin{abstract}
In psychotherapy interactions, the quality of a session is assessed by codifying the communicative behaviors of participants during the conversation through manual observation and annotation. Developing computational approaches for automated behavioral coding can reduce the burden on human coders and facilitate the objective evaluation of the intervention. In the real world, however, implementing such algorithms is associated with data sparsity challenges since privacy concerns lead to limited available in-domain data. In this paper, we leverage a publicly available conversation-based dataset and transfer knowledge to the low-resource behavioral coding task by performing an intermediate language model training via meta-learning. We introduce a task augmentation method to produce a large number of ``analogy tasks'' --- tasks similar to the target one --- and demonstrate that the proposed framework predicts target behaviors more accurately than all the other baseline models.
\end{abstract}

\section{Introduction}

Advances in spoken language processing techniques have improved the quality of life across several domains. One of the striking applications is automated behavioral coding in the fields of healthcare conversations such as psychotherapy. Behavioral coding is a procedure during which experts manually identify and annotate the participants' behaviors \citep{cooper2012apa}.
However, this process suffers from a high cost in terms of both time and human resources \citep{fairburn2011therapist}.
Building computational models for automated behavioral coding can significantly reduce the cost in time and provide scalable analytical insights into the interaction. A great amount of such work has been developed, including for addiction counseling \citep{tanana2016comparison, perez2017predicting, chen2019improving, flemotomos2022automated} and couples therapy \citep{li2016sparsely, tseng2016couples, biggiogera2021bert}. However, automated coding is associated with data sparsity due to the highly sensitive nature of the data and the costs of human annotation. Due to those reasons, {\it both} samples and labels of in-domain data are typically limited. This paper aims to train computational models for predicting behavior codes directly from psychotherapy utterances through classification tasks with limited in-domain data. 

Recently, substantial work has shown the success of universal language representation via pre-training context-rich language models on large corpora \cite{peters-etal-2018-deep, howard-ruder-2018-universal}. Particularly, BERT (Bidirectional Encoder Representations from Transformers) has achieved state-of-the-art performance in many natural language processing (NLP) tasks and provided strong baselines in low-resource scenarios \cite{devlin-etal-2019-bert}. However, these models rely on self-supervised pre-training on a large out-of-domain text corpus. In prior works, the data sparsity issue has also been addressed by introducing an intermediate task pre-training using some other high-resource dataset \cite{houlsby2019parameter, liu-etal-2019-linguistic, vu-etal-2020-exploring}. However, not all the source tasks yield positive gains. Sometimes the intermediate task might even lead to degradation due to the negative transfer \cite{pruksachatkun-etal-2020-intermediate, lange-etal-2021-share, poth2021pre}. To improve the chance of finding a good transfer source, we need to collect as many source tasks as possible. Another approach is meta-learning which aims to find optimal initialization for fine-tuning with limited target data \cite{gu-etal-2018-meta, dou-etal-2019-investigating, qian-yu-2019-domain}. This approach also calls for enough source tasks and is affected by any potential task dissimilarity \cite{jose2021information, zhou2021task}.

 The challenge we need to handle is that both utterances and assigned codes in psychotherapy interactions are domain-specific, making it difficult to leverage any open resource from a related domain. Considering that psychotherapy counseling takes place in a conversational setting, here we use a publicly available dataset --- Switchboard-DAMSL (SwDA) corpus \cite{stolcke2000dialogue} --- for the intermediate stages of knowledge transfer. 
 Unlike most previous meta-learning frameworks, which require auxiliary tasks from various datasets, our work uses only one dataset and produces the source tasks by a task augmentation procedure. The task augmentation framework evaluates the correlations between the source and target labels. It produces source tasks by choosing subsets of source labels whose classes are in one-to-one correspondence with the target classes. Using this strategy, we can generate a large number of source tasks similar to the target task and thus improve the performance of meta-learning. The experimental results show that incorporating our proposed task augmentation strategy into meta-learning enhances the classification accuracy of automated behavioral coding tasks and outperforms all the other baseline approaches.



\begin{table}[]
\begin{center}
\resizebox{0.45\textwidth}{!}{\begin{tabular}{|cccc|}
\hline
\multicolumn{1}{|c|}{\textbf{Code}} & \multicolumn{1}{c|}{\textbf{Description}} & \multicolumn{1}{c|}{\textbf{\#Train}} & \textbf{\#Test} \\ \hline
\multicolumn{4}{|c|}{Therapist Utterances}                                                                                                           \\ \hline
\multicolumn{1}{|c|}{FA}            & \multicolumn{1}{c|}{Facilitate}           & \multicolumn{1}{c|}{19397}            & 5838            \\ \hline
\multicolumn{1}{|c|}{GI}            & \multicolumn{1}{c|}{Giving information}   & \multicolumn{1}{c|}{17746}            & 5064            \\ \hline
\multicolumn{1}{|c|}{RES}           & \multicolumn{1}{c|}{Simple reflection}    & \multicolumn{1}{c|}{7236}             & 2137            \\ \hline
\multicolumn{1}{|c|}{REC}           & \multicolumn{1}{c|}{Complex reflection}   & \multicolumn{1}{c|}{4974}             & 1510            \\ \hline
\multicolumn{1}{|c|}{QUC}           & \multicolumn{1}{c|}{Closed question}      & \multicolumn{1}{c|}{6421}             & 1569            \\ \hline
\multicolumn{1}{|c|}{QUO}           & \multicolumn{1}{c|}{Open question}        & \multicolumn{1}{c|}{5011}             & 1475            \\ \hline
\multicolumn{1}{|c|}{MIA}           & \multicolumn{1}{c|}{MI adherent}          & \multicolumn{1}{c|}{4898}             & 1346            \\ \hline
\multicolumn{1}{|c|}{MIN}           & \multicolumn{1}{c|}{MI non-adherent}      & \multicolumn{1}{c|}{1358}             & 237             \\ \hline
\multicolumn{4}{|c|}{Patient Utterances}                                                                                                             \\ \hline
\multicolumn{1}{|c|}{FN}            & \multicolumn{1}{c|}{Follow/Neutral}       & \multicolumn{1}{c|}{56204}            & 15426           \\ \hline
\multicolumn{1}{|c|}{POS}           & \multicolumn{1}{c|}{Change talk}          & \multicolumn{1}{c|}{6146}             & 1737            \\ \hline
\multicolumn{1}{|c|}{NEG}           & \multicolumn{1}{c|}{Sustain talk}         & \multicolumn{1}{c|}{5121}             & 1407            \\ \hline
\end{tabular}}
\end{center}
\caption{Data statistics for behavior codes in
Motivational Interviewing psychotherapy.}\label{tab:MI_data}
\end{table}

\section{Dataset}
\label{sec:Dataset}

We use data from Motivational Interviewing (MI) sessions of alcohol and drug abuse problems \cite{baer2009agency, atkins2014scaling} for the target task. The corpus consists of 345 transcribed sessions with behavioral codes annotated at the utterance level according to the Motivational Interviewing Skill Code (MISC) manual \cite{houck2010motivational}. We split the data into training and testing sets with a roughly 80\%:20\% ratio across speakers, resulting in 276 training sessions and 67 testing sessions. The statistics of the data are shown in Table \ref{tab:MI_data}.

We perform the intermediate task with the SwDA dataset, which consists of telephone conversations with a dialogue act tag for each utterance. We concatenate the parts of an interrupted utterance together, following \citet{webb2005dialogue}, which results in 196K training utterances and 4K testing utterances. 
This dataset supports 42 distinct tags, with more details displayed in Appendix \ref{sec:swda}.
 
\section{Methodology}

\subsection{Task Augmentation via Label Clustering}
\label{sec:task_label}

We define a low resource target task on \mbox{$\mathcal{X} \times \mathcal{Y}$} and use \mbox{$x \in \mathcal{X}$} to denote data and \mbox{$y \in \mathcal{Y} = \{1,2,...,M\}$} to denote the target labels. We additionally assume a data-rich source task defined on \mbox{$\mathcal{X} \times \mathcal{Z}$} with samples $\{(x_1, z_1), (x_1, z_2),$ $...,(x_n, z_n)\}$ supported by a much larger label set denoted by $z \in \mathcal{Z} = \{1,2,...,N\}$, $N > M$. Our task augmentation procedure aims at producing numerous tasks similar to the target task---we will refer to those as the {\em ``analogy tasks''}. 

\begin{algorithm}[b]
\caption{Construction of Analogy Tasks}\label{alg:alg1}
\begin{algorithmic}
\State Initialize model parameters $\theta$; $K, M \in \mathbb{N}$.
\State Create empty label subsets: $C_1=\varnothing, C_2=\varnothing,...,C_M=\varnothing$.
\State Fine-tune BERT with in-domain samples to obtain the classifier $f$
\For{$i$ = 1 to $K$}
\For{$j$ = 1 to $M$}
\State For the target label $y=j$, select $z^* \in \mathcal{Z}$ by Equation(\ref{eq:1}) and (\ref{eq:2}), then add it to $C_j$
\State Remove the label $z^*$ from $\mathcal{Z}$
\EndFor
\EndFor
\State Select one label from $C_1, C_2,...,C_M$ to produce $M^K$ analogy tasks
\end{algorithmic}
\end{algorithm}

The high-level idea is to construct the tasks with class labels similar to the target ones. Thus we explore the relationships between $\mathcal{Y}$ and $\mathcal{Z}$. We initialize $M$ label subsets $C_1=\varnothing, C_2=\varnothing,\cdots,C_M=\varnothing$ to gather the source labels corresponding to $y=1, y=2,\cdots,y=M$, respectively.
In the first step, we fine-tune on the in-domain target data to achieve a dummy classifier $f$. Then, we feed the source samples into $f$ and obtain the predicted labels $\hat{Z} = \{f(x_1), f(x_2),\cdots,f(x_n)\}$. For any pair of a target label $y$ and a source label $z$, we define the similarity function $Sim(\cdot)$ expressed by Equation~(\ref{eq:1}). The value of $Sim(y,z)$ represents the proportion of the source samples within the class $z$, which are assigned the label $y$ by $f$. For any target label $y$, we determine the most similar label $z^*$ from the source data by Equation (\ref{eq:2}). Next, we apply Equations (\ref{eq:1}) and (\ref{eq:2}) to each of the target labels alternatively to cluster the source labels into the label subsets $C_1, C_2,\cdots,C_M$ with $|C_1|=|C_2|=\cdots=|C_M|=K$, where $K$ is the size of the label subsets. Finally, we generate the source tasks by selecting one label from each of the subsets, resulting in $M^K$ \emph{analogy tasks}. The details of this procedure are given in Algorithm \ref{alg:alg1}. 

\begin{figure}[]
  \centering
  \includegraphics[width=6.8cm]{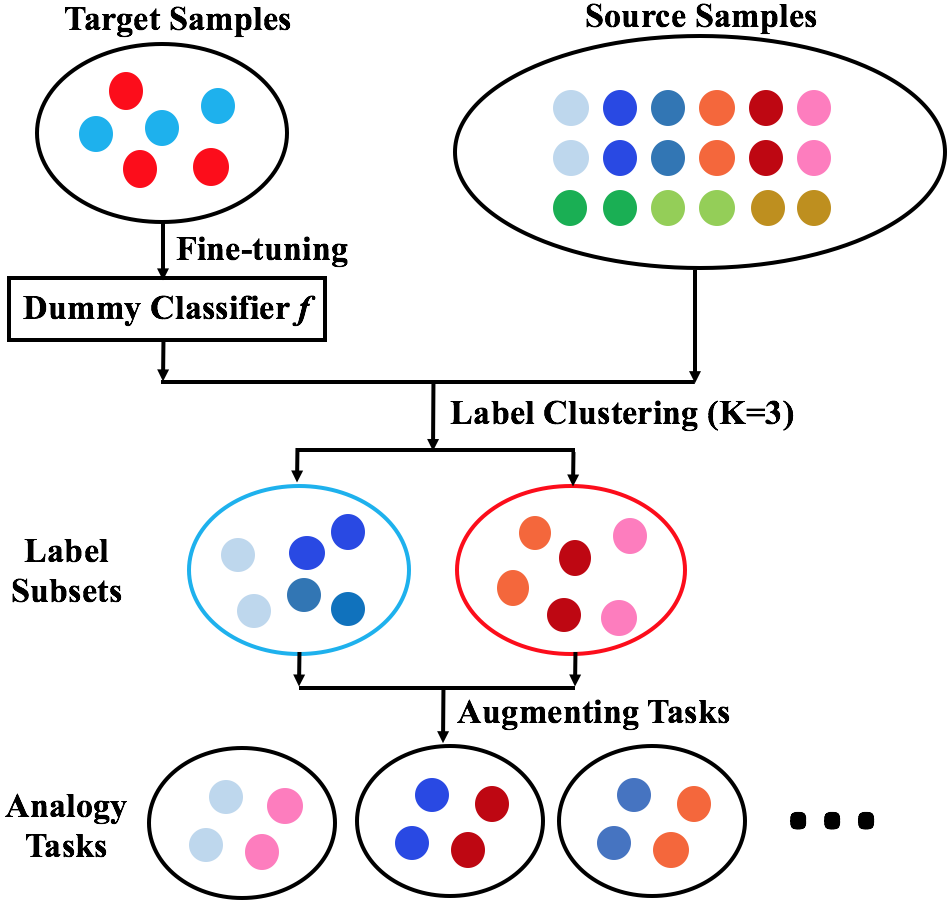}
  \caption{An example of task augmentation.}
  \label{fig:task_augment}
\end{figure}

\begin{align}
Sim(y,z) &= \frac{\sum_{k=1}^{n}\mathbbm{1}\{f(x_k)=y ,z_k=z\}}{\sum_{i=1}^{n}\mathbbm{1}\{z_k=z\}} \label{eq:1}\\
z^* &= \mathop{\arg \max}\limits_{z \in \mathcal{Z}} Sim(y,z) \label{eq:2}
\end{align}



Fig.~\ref{fig:task_augment} presents a task augmentation example from which we suppose the produced \emph{analogy tasks} can benefit meta-learning in three aspects: 1) the task similarity and knowledge transfer are improved; 2) the large number of the \emph{analogy tasks} increase the generalization which helps meta-learner find a commonly good model initialization; 3) the classification layers can be shared for all the tasks.

\subsection{Meta-learning with Analogy Tasks}
\label{sec:meta-augment}

After task augmentation, we apply an optimization-based meta-learning algorithm for intermediate training with the produced analogy tasks ${T_1, T_2,..., T_{M^K}}$. In particular, here we use Reptile, that has shown superior text classification results \citep{dou-etal-2019-investigating}. 
We denote this Reptile-based framework with task augmentation as \emph{Reptile-TA} and propose two task sampling methods:

\noindent\textbf{Uniform}, where we choose a task by uniformly selecting one source label from each label subset;


\noindent\textbf{PPTS}, where we choose an analogy task with the \textbf{p}robability \textbf{p}roportional to the \textbf{t}ask \textbf{s}ize to make the best use of instances (see Appendix \ref{sec:proof}).

We describe the training procedure in Algorithm~\ref{alg:alg2} where $\alpha$ and $\beta$ present the learning rate for the inner and outer loop, respectively, and $m$ denotes the update steps for the inner loop. 

\begin{algorithm}
\caption{Reptile with Analogy Tasks}\label{alg:alg2}
\begin{algorithmic}
\State Initialize model parameters $\theta$; $m \in \mathbb{N}$, $\alpha, \beta > 0$ 
\For{iteration in 1,2,...}
\State Sample a batch of analogy tasks $\{\tau_i\}$ based on one of the sampling methods we proposed.
\For{all $\tau_i$}
\State Compute $\theta_i^m$ by $m$ gradient descent steps with the learning rate $\alpha$.
\State Update $\theta$ = $\theta + \beta\frac{1}{|\{\tau_i\}|}\sum_{\tau_i}(\theta_i^m-\theta)$
\EndFor
\EndFor
\end{algorithmic}
\end{algorithm}
\setlength\textfloatsep{0.5\baselineskip plus 3pt minus 2pt}

\section{Experimental Results}

We adopt the MI dataset to perform two tasks: predicting the behavioral codes of the therapist and of the patient. We use SwDA as the source dataset for intermediate tasks to train the BERT model with Reptile. 
We set the number of sessions for both the training and validation sets to 1, 5, and 25 to simulate low-resource situations at different levels. We pick sessions randomly to make pairs of training and validation, and we repeat this 15 times. For each level of data sparsity, we report the averaged prediction results over 15 runs to reduce the effect of data variations. 

\subsection{Experimental Setup}

Our BERT model was implemented in PyTorch (version 1.3.1) and initialized with \textbf{BERT-base}\footnote{https://github.com/huggingface/pytorch-pretrained-BERT}. 
The model is trained using the Adam optimizer \citep{kingma2014adam} with a batch size of 64. In the Reptile stage, we set the learning rate to be $\alpha=$ 5e-5 for the inner loop and $\beta=$ 1e-5 for the outer loop and fix the inner update step $m$ to be 3 (Algorithm \ref{alg:alg2}). We pre-train the model for 4 epochs and sample 8 tasks in each step. In the fine-tuning stage, we select the learning rate from $\{$5e-6, 1e-5, 2e-5, 3e-5$\}$ and number of epochs from $\{$1, 3, 5, 10$\}$ with the lowest validation loss via grid search. To handle the class imbalance, we assign a weight for each class inversely proportional to its class frequency in the fine-tuning stage. In the meta-learning stage, we assign a weight for each sample inversely proportional to the frequency of the label subsets it belongs. The tasks are evaluated by the unweighted average recall (UAR).

\subsection{Baseline Methods}


\noindent\textbf{BERT:} We directly fine-tune BERT with the limited in-domain data.

\noindent\textbf{Pre-train-42:} We pre-train the intermediate task of BERT with the SwDA dataset using a 42-class classification task adopting its standard label tags.

\noindent\textbf{Pre-train-7:} We cluster the labels into simpler 7 tags, as described by \citet{shriberg1998can}, and pre-train the intermediate task of BERT with the SwDA dataset using a 7-class classification task.

\begin{table}[]
\begin{center}
\resizebox{0.48\textwidth}{!}{
\begin{tabular}{cccc}
\hline
\multirow{2}{*}{Approach} & \multicolumn{3}{c}{Nb. Training Sessions}                               \\ \cline{2-4} 
                          & \multicolumn{1}{c}{1}              & \multicolumn{1}{c}{5}     & 25    \\ \hline
BERT                      & \multicolumn{1}{c}{0.512}          & \multicolumn{1}{c}{0.577} & 0.626 \\ 
Pre-train-42              & \multicolumn{1}{c}{0.528}          & \multicolumn{1}{c}{0.584} & 0.630 \\
Pre-train-7               & \multicolumn{1}{c}{0.543}          & \multicolumn{1}{c}{0.592} & 0.638 \\
Pre-train-LC-Shared       & \multicolumn{1}{c}{0.533}          & \multicolumn{1}{c}{0.584} & 0.633 \\
Pre-train-LC-Unshared     & \multicolumn{1}{c}{0.552}          & \multicolumn{1}{c}{0.597} & 0.643 \\
\hdashline
{Reptile-TA-Uniform}         & \multicolumn{1}{c}{0.555}          & \multicolumn{1}{c}{0.601} & 0.646 \\
{Reptile-TA-PPTS}         & \multicolumn{1}{c}{\textbf{0.574}} & \multicolumn{1}{c}{\textbf{0.618}} & \textbf{0.660} \\
 \hline
\end{tabular}}
\end{center}
\caption{UARs on predicting therapist's codes.}\label{tab:T_result}
\end{table}

To explore the effect of label clustering, we propose two more baseline approaches:

\noindent\textbf{Pre-train-LC-Shared:} After label clustering, we pre-train the model by classifying samples into the label subsets they belong to as in Fig.~\ref{fig:task_augment}. The classification layer is shared between pre-training and fine-tuning stage.

\noindent\textbf{Pre-train-LC-Unshared:} The setup is the same as in \textbf{Pre-train-LC-Shared}, but the classification layer is randomly initialized for fine-tuning.

\subsection{Results}

The results of different algorithms for predicting therapist's and patient's codes are presented in Tables~\ref{tab:T_result} and~\ref{tab:P_result}. For the therapist-related tasks, both \emph{Pre-train-42} and \emph{Pre-train-7} outperform fine-tuning BERT directly because some of the therapist's codes (i.e., ``Open Question'' or ``Closed Question'') are similar in function to dialog acts such as ``Open Question'' and ``Yes-No Question''. The \emph{Pre-train-7} groups the source labels in a reasonable way, making the source task closer to the target task and achieving better performance than \emph{Pre-train-42}. However, both failed to improve the accuracy of predicting patient behavior since the codes reflect whether the patient shows a motivation to change their behavior and thus do not have evident similarities to these dialogue acts. The results of \emph{Pre-train-LC-Unshared} are better when compared to direct fine-tuning and regular pre-training. The greater improvement in the patient's task indicates that gathering the source labels similar to target labels is effective. However, sharing the classification layer when fine-tuning degrades the performance in the task of therapists. This drop is because the pre-trained models do not provide a good initialization, and thus, when we fine-tune BERT, it becomes difficult to escape from local minima.

The results under the dashed line in Tables \ref{tab:T_result} and \ref{tab:P_result} are for our proposed framework, where we set the size of label subsets $K$ to be 3 and 8 for the therapist's task and patient's task, respectively. The outcomes show that our framework with task augmentation performs better than the baseline approaches. We further compare the performance using the different task sampling strategies proposed in Section~\ref{sec:meta-augment}, and the results demonstrate that \emph{PPTS} is superior to \emph{Uniform} achieving significantly better UAR scores than any other approaches at ($p < 0.05$) based on Student’s t-test.

\begin{table}[]
\begin{center}
\resizebox{0.48\textwidth}{!}{
\begin{tabular}{cccc}
\hline
\multirow{2}{*}{Approach} & \multicolumn{3}{c}{Nb. Training Sessions}                               \\ \cline{2-4} 
                          & \multicolumn{1}{c}{1}              & \multicolumn{1}{c}{5}     & 25    \\ \hline
BERT                      & \multicolumn{1}{c}{0.408}          & \multicolumn{1}{c}{0.469} & 0.528 \\ 
Pre-train-42              & \multicolumn{1}{c}{0.407}          & \multicolumn{1}{c}{0.463} & 0.523 \\
Pre-train-7               & \multicolumn{1}{c}{0.410}          & \multicolumn{1}{c}{0.466} & 0.529 \\
Pre-train-LC-Shared       & \multicolumn{1}{c}{0.445}          & \multicolumn{1}{c}{0.497} & 0.545 \\
Pre-train-LC-Unshared     & \multicolumn{1}{c}{0.446}          & \multicolumn{1}{c}{0.499} & 0.545 \\
\hdashline
{Reptile-TA-Uniform}         & \multicolumn{1}{c}{0.448}          & \multicolumn{1}{c}{0.499} & 0.547 \\
{Reptile-TA-PPTS}         & \multicolumn{1}{c}{\textbf{0.461}} & \multicolumn{1}{c}{\textbf{0.511}} & \textbf{0.555} \\
 \hline
\end{tabular}}
\end{center}
\vspace{-0.2cm}
\caption{UARs on predicting patient's codes.}\label{tab:P_result}
\vspace{-0.2cm}
\end{table}

\subsection{Effect of the Size of Label Subsets}

We test the effect of $K$ using \emph{Pre-train-LC-Unshared} and
\emph{Reptile-TA-PPTS} with 5 training sessions. From the results in Tables \ref{tab:T_K} and \ref{tab:P_K} we find that an optimal $K$ should be neither too small nor too big. When $K$ is small, we utilize too little source data. A bigger value of $K$ leads to a larger number of samples and augmented tasks. However, at the same time, it weakens the task similarity.

\begin{table}[]
\begin{center}
\resizebox{0.48\textwidth}{!}{
\begin{tabular}{ccccc}
\hline
\multirow{2}{*}{Approach} & \multicolumn{4}{c}{Size of label subset K}                                                                    \\ \cline{2-5} 
                          & \multicolumn{1}{c}{2}              & \multicolumn{1}{c}{3}              & \multicolumn{1}{c}{4}     & 5     \\ \hline
Pre-train-LC-Unshared     & \multicolumn{1}{c}{0.585}          & \multicolumn{1}{c}{\textbf{0.597}} & \multicolumn{1}{c}{0.596} & 0.589 \\
Reptile-TA-PPTS         & \multicolumn{1}{c}{0.603} & \multicolumn{1}{c}{\textbf{0.618}} & \multicolumn{1}{c}{0.615} & 0.608 \\ \hline
\end{tabular}}
\end{center}
\vspace{-0.2cm}
\caption{Effect of the size of label subset $K$, 8-way classification tasks of therapist.}\label{tab:T_K}
\vspace{0.2cm}
\end{table}

\vspace{0.2cm}
\begin{table}[]
\begin{center}
\resizebox{0.48\textwidth}{!}{
\begin{tabular}{ccccc}
\hline
\multirow{2}{*}{Approach} & \multicolumn{4}{c}{Size of label subset K}                                                                    \\ \cline{2-5} 
                          & \multicolumn{1}{c}{2}              & \multicolumn{1}{c}{5}              & \multicolumn{1}{c}{8}     & 11     \\ \hline
Pre-train-LC-Unshared     & \multicolumn{1}{c}{0.477}          & \multicolumn{1}{c}{0.492} & \multicolumn{1}{c}{\textbf{0.499}} & 0.493 \\
Reptile-TA-PPTS         & \multicolumn{1}{c}{0.488} & \multicolumn{1}{c}{0.501} & \multicolumn{1}{c}{\textbf{0.511}} & 0.504 \\ \hline
\end{tabular}}
\end{center}
\caption{Effect of the size of label subset $K$, 3-way classification tasks of patient.}\label{tab:P_K}
\end{table}

\section{Conclusion and Future Work}

This paper leveraged publicly available datasets to build computational models for predicting behavioral codes in psychotherapy conversations with limited samples. We employed a meta-learning framework with task augmentation based on the idea of analogy tasks to address the data limitation problem. We performed experiments at different sparsity levels and showed improvement over baseline methods. Besides, we discussed two task sampling strategies and the effect of a hyper-parameters in our framework. In the future, we plan to leverage contextual utterances into our algorithm and generalize our approach to the natural language understanding task in other fields. A more formal approach to find an optimal match between classes from different domains (i.e., labels of conversational descriptors) is also a topic of our ongoing research.

\section{Limitations}

As an initial stage in an ongoing effort, our work has several limitations. First, we only leverage a single open dataset for the intermediate task. There are other conversation-based corpora with utterance-level labels that we have not explored yet, such as Persuasion For Good Corpus \citep{wang-etal-2019-persuasion} and DailyDialog Corpus \citep{li-etal-2017-dailydialog}. Second, we adopted the \textbf{BERT-base} as the language model throughout all the experiments ignoring domain adaptation. For example, we can perform domain-adaptive pre-training with a publicly available general psychotherapy corpus \citep{imel2015computational}. In our framework, we force the size of the label subsets to be the same in the label clustering stage, which might be sub-optimal. A more sophisticated clustering algorithm is needed. Besides, the intermediate task can introduce biases into the target, which calls for more discussion.

\section{Ethical Considerations}
As a research focused on psychotherapy and automated behavioral coding using speech and language processing techniques, it is necessary to review the ethical implications of this work. 

Given the sensitive nature of the data, the primary ethical issue is the privacy of all the participating individuals - both patients and therapists. Informed consent was employed to make sure the recording is permitted by the participants, in adherence to professional guidelines \citep{american2002ethical}. All the researchers involved in the study are trained and certifies on human subject data research, and all the data are stored in dedicated secure machines with restricted access. It was guaranteed that these data will not be shared with anyone who is not involved in the study. The current study is governed by restrictions imposed by the relevant Institutional Review Board (IRB).


\bibliography{anthology,custom}
\bibliographystyle{acl_natbib}

\clearpage
\appendix

\section*{Appendix}

\section{The SwDA Dataset}
\label{sec:swda}

This section demonstrates the dialogue acts distributions of the SwDA dataset.  The statistics for the 42-tag scheme and the simpler 7-tag scheme are presented in Tables \ref{tab:swda_42}
and \ref{tab:swda_7}, respectively. 

\section{A proposition for the Sampling Strategy PPTS}
\label{sec:proof}

\begin{proposition}
\label{thm:thm2}
If we adopt the sampling strategy PPTS, then every unique instance within the label subsets has the same chance of being picked.
\end{proposition}

\noindent\emph{Proof.} Define the label subsets after label clustering $C_1=\{c_1^1, c_1^2,..., c_1^K\}$, $C_2=\{c_2^1, c_2^2,..., c_2^K\}$,..., $C_M=\{c_M^1, c_M^2,..., c_M^K\}$.\\

Let $x$ be an arbitrary instance with label which is contained in the label subset $C_i$, $1 \leq i \leq M$.

Consider the following process: 1) sample an analogy task with the probability proportional to the task size; 2) randomly sample an instance from the selected task.

We compute the probability of the picked instance to be $x$ by

\begin{equation} \label{eq:3}
\begin{split}
P(x) &= \sum_{T} P(T) \cdot P(x|T) \\
&= \sum_{T} \frac{|T|}{\sum_{T}|T|} \cdot \frac{1}{|T|} \cdot \mathbbm{1}\{x \in T\} \\
&=  \sum_{T} \frac{\mathbbm{1}\{x \in T\}}{\sum_{T}|T|}
\end{split}
\end{equation}

where $T$ denotes any analogy task. Consider that an arbitrary label $c$ can be enrolled in exact $K^{M-1}$ analogy tasks. Equation \ref{eq:3} can be rewritten as  

\begin{equation} \label{eq:4}
\begin{split}
P(x) &=\sum_{T} \frac{K^{M-1}}{K^{M-1}\sum_i^M\sum_j^K |c_i^j|} \\
&= \frac{1}{\sum_i^M\sum_j^K |c_i^j|}
\end{split}
\end{equation}

The probability is irrelevant to the label and thus the same for every instance $x$. Please note that the proposition will not hold if the sizes of the label subsets $C_i$ are different.

\section{Examples of Label Subsets}
\label{sec:eg_lc}

This part shows examples of label clustering results for predicting therapist's codes and patient's codes with five in-domain training sessions. Table \ref{tab:subset_T} and \ref{tab:subset_P} present the produced label subsets which achieve the median performance among 15 runs of \emph{Reptile-TA-PPTS}.

\begin{table*}[]
\begin{center}
\resizebox{0.85\textwidth}{!}{
\begin{tabular}{|c|c|c|c|}
\hline
Dialogue Act                 & Utterances (count) & Dialogue Act                 & Utterances (count) \\ \hline
statement-non-opinion        & 74k                & collaborative completion     & 0.7k               \\ 
backchannel                  & 38k                & repeat-phrase                & 0.7k               \\ 
statement-opinion            & 26k                & open question                & 0.6k               \\ 
abandoned/uninterpretable    & 15k                & rhetorical questions         & 0.6k               \\ 
agree/accept                 & 11k                & hold-before-answer/agreement & 0.5k               \\ 
appreciation                 & 4.7k               & reject                       & 0.3k               \\ 
yes-no-question              & 4.7k               & negative non-no answers      & 0.3k               \\ 
non-verbal                   & 3.6k               & signal-non-understanding     & 0.3k               \\ 
yes answers                  & 3k                 & other answer                 & 0.3k               \\ 
Conventional-closing         & 2.6k               & conventional-opening         & 0.2k               \\ 
wh-question                  & 1.9k               & or-clause                    & 0.2k               \\ 
no answers                   & 1.4k               & dispreferred answers         & 0.2k               \\ 
response acknowledgement     & 1.3k               & 3rd-party-talk               & 0.1k               \\ 
hedge                        & 1.2k               & offers, options commits      & 0.1k               \\ 
declarative yes-no-question  & 1.2k               & self-talk                    & 0.1k               \\ 
backchannel in question form & 1k                 & downplayer                   & 0.1k               \\ 
quotation                    & 0.9k               & maybe/accept-part            & 0.1k               \\ 
summarize/reformulate        & 0.9k               & tag-question                 & 0.1k               \\ 
other                        & 0.9k               & declarative wh-question      & 0.1k               \\ 
affirmative non-yes answers  & 0.8k               & apology                      & 0.1k               \\ 
action-directive             & 0.7k               & thinking                     & 0.1k               \\ \hline
\end{tabular}}
\end{center}
\caption{Statistics describing the SwDA datasets for the 42 tags scheme.}\label{tab:swda_42}
\end{table*}

\begin{table*}[]
\begin{center}
\resizebox{0.30\textwidth}{!}{
\begin{tabular}{|c|c|}
\hline
Dialogue Act & Utterances (count) \\ \hline
statement    & 100k               \\
backchannel  & 38k                \\
question     & 8.6k               \\
agreement    & 11k                \\
appreciation & 4.7k               \\
incomplete   & 15k                \\
other        & 23k                \\ \hline
\end{tabular}}
\end{center}
\caption{Statistics describing the SwDA datasets for the 7 tags scheme \cite{shriberg1998can}.}\label{tab:swda_7}
\end{table*}

\begin{table*}[]
\begin{center}
\resizebox{0.75\textwidth}{!}{
\begin{tabular}{|c|c|}
\hline
Behavioral Code    & Clustered Similar Labels                                        \\ \hline
Facilitate         & backchannel, yes answer, no answer                              \\ \hline
Giving Information & statement-opinion, statement-non-opinion, dispreferred-answers  \\ \hline
Simple Reflection  & quotation, declarative yes-no-question, declarative wh-question \\ \hline
Complex Reflection & non-verbal, hedge, summarize/reformulate                        \\ \hline
Closed Question    & yes-no-question, or-clause, tag-question                        \\ \hline
Open Question      & wh-question, open-question, self-talk                           \\ \hline
MI adherent        & appreciation, downplayer, thanking                              \\ \hline
MI non-adherent    & action-directive, offers/options commits, 3rd-party-talk        \\ \hline
\end{tabular}}
\end{center}
\caption{Label clustering results for therapist’s codes
.}\label{tab:subset_T}
\end{table*}

\begin{table*}[]
\begin{center}
\resizebox{0.75\textwidth}{!}{
\begin{tabular}{|c|c|}
\hline
Behavioral Code         & Clustered Similar Labels                                                                                                                                                                                              \\ \hline
Follow/Neutral          & \begin{tabular}[c]{@{}c@{}}backchannel, no answer, non-verbal, yes answer,\\ response acknowledgement, tag-question, repeat-phrase,\\ backchannel in question form\end{tabular}                                       \\ \hline
\begin{tabular}[c]{@{}c@{}}Change Talk\\ (positive)\end{tabular}  & \begin{tabular}[c]{@{}c@{}}quotation, declarative, yes-no-question, offers/options commits,\\ statement-opinion, declarative wh-question, rhetorical-questions, \\ 3rd-party-talk,  yes-no-question\end{tabular}      \\ \hline
\begin{tabular}[c]{@{}c@{}}Sustain Talk\\ (negative)\end{tabular}  & \begin{tabular}[c]{@{}c@{}}statement-non-opinion,  collaborative completion, hedge,\\ action-directive, other answers,  dispreferred answers,\\ declarative yes-no-question, affirmative non-yes answers\end{tabular} \\ \hline
\end{tabular}}
\end{center}
\caption{Label clustering results for patient’s codes
.}\label{tab:subset_P}
\end{table*}

\end{document}